\documentclass{article}
\usepackage{spconf,amsmath,graphicx,hyperref}
\usepackage{tabularx,multirow}
\usepackage{graphicx}
\usepackage{booktabs}

\title{Robust Photoplethysmography Signal Denoising via Mamba Networks}
\name{I Chiu$^{1}$,
Yu-Tung Liu$^{2}$,
Kuan-Chen Wang$^{3}$,
Hung-Yu Wei$^{4}$,
and Yu Tsao$^{2}$
}

\address{
\small $^{1}$Taiwan International Graduate Program in Artificial Intelligence of Things,\\
\small Academia Sinica and National Taiwan University, Taipei, Taiwan\\
\small $^{2}$Research Center for Information Technology Innovation, Academia Sinica, Taipei, Taiwan\\
\small $^{3}$Graduate Institute of Communication Engineering, National Taiwan University, Taipei, Taiwan\\
\small $^{4}$Department of Electrical Engineering, National Taiwan University, Taipei, Taiwan\\
\small Email: d13949002@ntu.edu.tw, yu.tsao@citi.sinica.edu.tw
}

\begin{document}
\ninept
\maketitle
\begin{abstract}
Photoplethysmography (PPG) is widely used in wearable health monitoring, but its reliability is often degraded by noise and motion artifacts, limiting downstream applications such as heart rate (HR) estimation. This paper presents a deep learning framework for PPG denoising with an emphasis on preserving physiological information. In this framework, we propose DPNet, a Mamba-based denoising backbone designed for effective temporal modeling. To further enhance denoising performance, the framework also incorporates a scale-invariant signal-to-distortion ratio (SI-SDR) loss to promote waveform fidelity and an auxiliary HR predictor (HRP) that provides physiological consistency through HR-based supervision. Experiments on the BIDMC dataset show that our method achieves strong robustness against both synthetic noise and real-world motion artifacts, outperforming conventional filtering and existing neural models. Our method can effectively restore PPG signals while maintaining HR accuracy, highlighting the complementary roles of SI-SDR loss and HR-guided supervision. These results demonstrate the potential of our approach for practical deployment in wearable healthcare systems.
\end{abstract}
\section{Introduction}
\label{sec:introduction}
A photoplethysmography (PPG) recording measures the peripheral pulse by detecting variations in light transmission or reflection through capillaries. It is widely deployed in wearable devices, such as smart watches, rings, and fitness trackers, to noninvasively monitor vital signs, including heart rate (HR), oxygen saturation, and blood pressure~\cite{kim2023photoplethysmography}. For reliable health monitoring and accurate interpretation of physiological conditions, acquiring high‐fidelity PPG signals is essential.

However, obtaining clean PPG recordings in real‐world settings poses significant challenges. Factors such as sensor, skin motion, motion‐induced changes in blood flow, and ambient light fluctuations introduce artifacts that distort the waveform and degrade downstream performance~\cite{charlton20232023, hartmann2019toward, biswas2019cornet}. Because these artifacts can span a wide frequency band or exhibit nonstationary behavior, traditional signal‐processing techniques often struggle to remove them effectively~\cite{mishra2020survey}. Consequently, developing robust PPG‐denoising algorithms remains a critical requirement for dependable wearable health applications.

Recent approaches to PPG denoising have primarily employed deep learning architectures to reconstruct clean waveforms. One common strategy is the use of bidirectional long short-term memory (BLSTM) networks~\cite{lee2018bidirectional}, where stacked BLSTM layers process noisy PPG segments and leverage bidirectional context to recover the underlying clean signals. However, BLSTMs suffer from their recurrent design, which restricts parallelism and results in slow inference. In addition, they have limited capacity to capture long-range dependencies. Transformer-based models~\cite{vaswani2017attention} offer an alternative, excelling at capturing global temporal context and showing strong potential for PPG denoising. Yet, their quadratic complexity with respect to sequence length results in substantial computational cost~\cite{kwon2022preeminently, chen2024adapting}. Meanwhile, limited labeled PPG data can make Transformers prone to underfitting, and even well-trained models remain difficult to deploy in real time on resource-constrained wearable devices~\cite{bousefsaf20193d}. In addition, most existing approaches focus primarily on waveform reconstruction using pointwise objective functions (e.g., L1 or L2), with limited attention to downstream tasks such as HR estimation, where subtle distortions may undermine clinical reliability.

To address both the modeling inefficiencies and the lack of physiological awareness in prior works, we introduce a novel deep learning framework for PPG denoising that builds upon the Mamba ~\cite{gu2023mamba} selective state‐space model (SSM). Mamba is an emerging sequence-modeling architecture that maintains linear-time complexity while effectively capturing long-range temporal dependencies.
Mamba‐based models are well suited to learn the quasi‐periodic patterns inherent in signals, even under highly noisy, dynamic conditions~\cite{chen2025denoisemamba}.
Beyond leveraging the Mamba architecture, this study incorporates an auxiliary HR predictor (HRP) that estimates beats-per-minute (BPM) from PPG segments. The HRP provides an additional training loss that guides denoising toward physiological consistency, improving both waveform fidelity and HR estimation accuracy.

\begin{figure*}[t!]
    \centering
    \includegraphics[width=1.8\columnwidth]{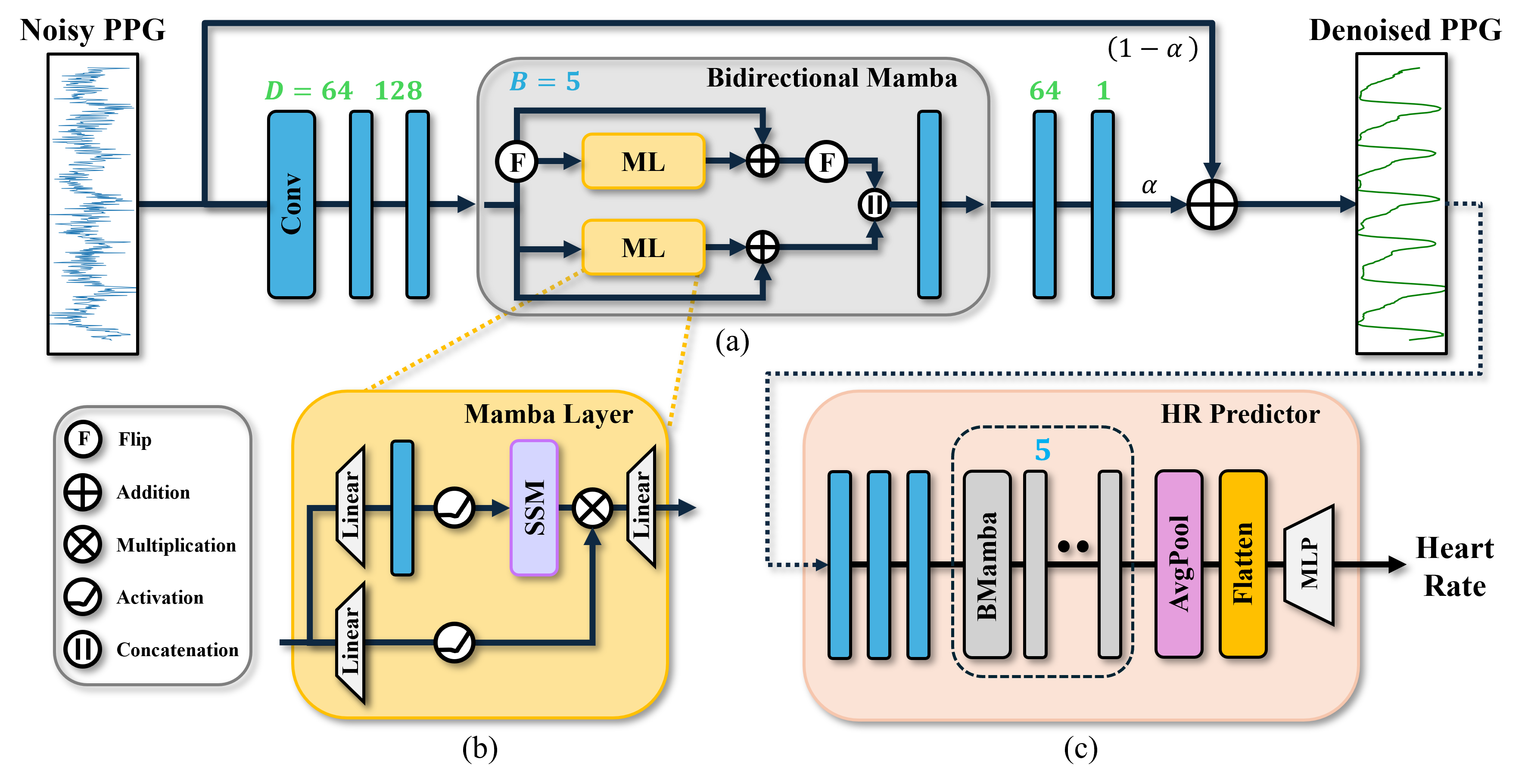}
    \caption{Architectures of (a) DPNet, (b) Mamba layer, and (c) HRP.}
    \label{fig:model}
\end{figure*}

To evaluate the proposed method, we design an experimental setup that considers both synthetic noise and real‐world motion artifacts. For clean PPG references, we adopt the Beth Israel Deaconess Medical Centre (BIDMC) dataset~\cite{pimentel2016toward}, which contains recordings collected in a controlled environment. Synthetic noise is added to the clean signals following the mixture procedures used in the previous studies~\cite{lee2018bidirectional, kwon2022preeminently}.
On the other hand, we extract motion artifacts from the WristPPG dataset~\cite{jarchi2016description}, where PPG signals are collected during everyday activities, introducing realistic motion-induced noise.
To further assess downstream utility in addition to waveform fidelity, we incorporate HR estimation as a task-specific metric. The same HR-extraction algorithm is applied to denoised outputs from the proposed framework, the conventional filtering~\cite{shenoi2005introduction}, and from baseline methods~\cite{lee2018bidirectional, ahmed2023deep}, enabling a consistent comparison of both signal fidelity and physiological reliability.



\section{Related works}
\subsection{PPG Denoising}
Early deep learning approaches for PPG denoising employed recurrent neural network (RNN) or convolutional neural network  (CNN) architectures. Lee et al. propose a Bidirectional Recurrent Auto-Encoder (BRDAE) that reconstructs clean waveforms from synthetically contaminated inputs using a bi-directional RNN~\cite{lee2018bidirectional}. Although RNNs can capture temporal dependencies, their sequential nature limits parallelism and computational efficiency. They also struggle with long-range dependencies, leading to unstable training and high memory costs.
Ahmed et al., on the other hand, propose a hybrid denoising framework called FWT-FFNN~\cite{ahmed2023deep}. In this method, the noisy PPG signal is first decomposed into multi-resolution subbands using the Fast Wavelet Transform (FWT), which captures both time- and frequency-domain characteristics. These wavelet coefficients are then fed into a feedforward neural network (FFNN) that learns to suppress noise while retaining physiologically relevant features. FWT-FFNN leverages the complementary strengths of wavelet decomposition and neural networks to achieve effective PPG denoising. In both methods, they mainly focus on waveform reconstruction and do not explicitly consider downstream tasks. In contrast, our approach is designed to enhance denoising performance while simultaneously improving physiological analysis.


\subsection{Mamba}
Beyond those denoising-specific designs, recent developments in sequence modeling highlight SSMs as a new paradigm for capturing long-range dependencies with improved efficiency.
Currently, the Mamba~\cite{gu2023mamba} model has been proposed as an alternative to Transformer architectures.
Mamba processes sequential data by evolving a latent state over time. At each step, the hidden state $\boldsymbol{h}$ is updated based on both the previous state and the current input $\boldsymbol{x}$. The output $\boldsymbol{y}$ is then generated through a projection:
\begin{align}
h_{n}&=\boldsymbol{\Bar{A}}h_{n-1}+\boldsymbol{\Bar{B}}x_{n},\\
y_{n}&=\boldsymbol{C}h_{n},
\end{align}
where $\boldsymbol{\Bar{A}}$, $\boldsymbol{\Bar{B}}$, and $\boldsymbol{C}$ are trainable matrices defining the state dynamics and interactions between input and output.
Mamba's unique design enables it to efficiently capture long-range dependencies. In various machine learning tasks, including computer vision~\cite{liu2024vmamba, li2024videomamba}, natural language processing~\cite{gu2023mamba, lieber2024jamba}, speech signal processing~\cite {li2024spmamba, chao2024investigation}, and biomedical signal processing~\cite{hung2024mecg, liu2025msemg, lin2025msecg}, Mamba has demonstrated performance comparable to Transformers while significantly reducing computational requirements. In this study, we explore the use of Mamba within a PPG denoising framework, marking the first application of such architectures in this domain. In addition, we introduce an auxiliary Mamba-based HR predictor, which explicitly improves downstream physiological analysis.

\section{Proposed method}\label{sec:proposed}
\subsection{Architecture}
The architecture of the proposed denoising model, DPNet, is illustrated in Fig.~\ref{fig:model}(a). The noisy PPG signal is first passed through three convolutional layers to extract local features, where $D$ denotes the number of output channels. These features are then fed into a sequence of bidirectional Mamba (BMamba) blocks~\cite{gu2023mamba}, which enable the model to capture long-range temporal dependencies in both forward and backward directions. The internal structure of the Mamba block is shown in Fig.~\ref{fig:model}(b). Afterward, two additional convolutional layers reduce the feature dimension to a single channel. Finally, the denoised signal is obtained through an element-wise weighted addition of the input and the network’s output, where the learnable parameter $\alpha$ adaptively balances the contributions of the preserved input and the reconstructed features. By integrating convolutional layers with Mamba, the model effectively captures both local details and global dependencies, enhancing robustness against diverse noise patterns.

To further guide this denoising process, we incorporate an auxiliary HR predictor (HRP), as illustrated in Fig.~\ref{fig:model}(c). This model is designed solely to estimate the HR from denoised PPG signals and provide an auxiliary loss during training. It consists of three convolutional layers, five BMamba blocks for temporal modeling, and a lightweight regression head formed by average pooling and a multilayer perceptron (MLP) to predict HR. The HRP provides additional supervision, encouraging DPNet to retain physiologically meaningful information in the reconstructed signals.

\section{Experiments}\label{sec:experiment}
\subsection{Datasets}\label{ssec:dataset}
In this study, we employ the BIDMC PPG and Respiration Dataset as the source of clean PPG signals~\cite{pimentel2016toward}. This dataset is a subset of the MIMIC-II Matched Waveform Database~\cite{saeed2011multiparameter}, containing simultaneously recorded physiological waveforms sampled at 125 Hz and physiological parameters sampled at 1 Hz. It includes key signals such as PPG, respiration, and derived parameters such as HR and respiratory rate (RR). Each subject's recording is 8 minutes in length, and the dataset comprises a total of 53 subjects.

In addition, we use the Wrist PPG During Exercise dataset~\cite{jarchi2016description} as a source of real-world motion artifacts. This dataset contains 18 wrist-worn PPG signals recorded from eight subjects during physical activities such as running and cycling, along with accelerometer and gyroscope measurements to capture motion dynamics. Each recording is approximately 10 minutes long and sampled at 256 Hz under realistic activity conditions, providing a reliable basis for evaluating denoising performance in the presence of motion-induced artifacts.


\begin{table*}[t]
\centering
\renewcommand\arraystretch{1.2}
\small
\caption{Quantitative performance comparison of different methods on the PPG denoising task.}
\label{tab:comparison}
\begin{tabular}{lcccc}
\toprule
Method & MSE ($\times 10^{-3}$) $\downarrow$ & CoS $\uparrow$ & SNR$_{imp}$ (dB) $\uparrow$ & HR-MAE $\downarrow$ \\
\midrule
Noisy  & 324.029$\pm$723.622 & 0.726$\pm$0.272 & - & 109.410$\pm$324.619\\ \hline
BP filters~\cite{shenoi2005introduction}  & 29.531$\pm$25.088 & 0.823$\pm$0.144 & -0.206$\pm$4.744 & 5.168$\pm$16.556\\
BRDAE~\cite{lee2018bidirectional}  & 19.229$\pm$18.578 & 0.881$\pm$0.126 & 2.054$\pm$3.574 & 2.896$\pm$21.060\\
FWT-FFNN~\cite{ahmed2023deep} & 31.562$\pm$36.412 & 0.802$\pm$0.237 & 1.517$\pm$2.726 & 11.510$\pm$36.174\\
DPNet (The proposed) & \textbf{6.663}$\pm$\textbf{9.845} & \textbf{0.961}$\pm$\textbf{0.069} & \textbf{8.323}$\pm$\textbf{4.789} & \textbf{1.025}$\pm$\textbf{4.869}\\
\bottomrule
\multicolumn{5}{l}{\textbf{Bold} represents the best performance.}
\end{tabular}
\end{table*}


\subsection{Data pre-processing and preparation}
\label{ssec:preprocessing}
The PPG signals are segmented into overlapping 6-second windows with a 4-second overlap. Each segment is analyzed with the Heartpy python package~\cite{van2019heartpy} and retained only if it meets the quality criteria: (40$<$bpm$<$150), (400$<$ibi$<$2000), (rmssd$<$100), (sd1sd2$<$6.0). Segments satisfied these conditions are used as ground truth for the denoising task, with their corresponding BPM values for the HR prediction task. This retained dataset is divided into training, validation, and test sets with an 8:1:1 ratio.

To simulate noisy conditions, we contaminate the clean PPG signals with synthetic noise and real-world motion artifacts. For synthetic noise, we introduce seven disturbances commonly observed in practice, including Gaussian noise, sloping baseline, saturation distortion, Poisson noise, salt-and-pepper noise, speckle noise, and uniform noise. 
Each noise type is appropriately scaled and superimposed onto the clean PPG waveforms, following previous works~\cite{lee2018bidirectional, kwon2022preeminently}.
Motion artifacts are derived from the Wrist PPG During Exercise dataset~\cite{jarchi2016description}. The raw signals are downsampled to 125 Hz to match the clean PPG. A motion segment of equal length is randomly aligned with the clean PPG, and a randomly chosen portion (0–100\% of its length) is blended by averaging amplitudes. This ensures that the resulting noisy signals retain both the underlying PPG morphology and the motion-induced disturbances.

\subsection{Evaluation metrics}
\label{ssec:metric}
Four metrics are adopted to evaluate all methods' signal reconstruction ability.
In every metric below, $\boldsymbol{g}$ denotes the ground truth (GT) PPG, $\boldsymbol{n}$ denotes the contaminated PPG, and $\boldsymbol{d}$ denotes the denoised PPG segments.

$\bullet$ Mean Squared Error (MSE): Quantifies the average squared difference between $\boldsymbol{g}$ and $\boldsymbol{d}$. It is defined as:
\begin{equation}
\text{MSE} = \frac{1}{N} \sum_{i=1}^{N} (\boldsymbol{g}[i] - \boldsymbol{d}[i])^2,
\end{equation}
where \( N \) is the number of samples in the signal. Lower MSE values indicate higher reconstruction accuracy.

$\bullet$ Cosine Similarity (CoS): Reflects how similar the directions of $\boldsymbol{g}$ and $\boldsymbol{d}$ are in a multi-dimensional space:
\begin{equation}
    \mathrm{CoS} = 
    \dfrac{\boldsymbol{g} \cdot \boldsymbol{d}}
    {\| \boldsymbol{g} \| \| \boldsymbol{d} \|},
\end{equation}
where $\| \cdot \|$ is the Euclidean norms of a signal. The CoS is useful for evaluating the similarity of waveform shape.

$\bullet$ Signal-to-Noise Ratio improvement (SNR$_{imp}$): Measures the relative improvement in SNR achieved by the denoised signal compared to the noisy input:
\begin{equation}
\text{SNR}_{imp} = \text{SNR}(\boldsymbol{d}, \boldsymbol{g}) - \text{SNR}(\boldsymbol{n}, \boldsymbol{g}).
\end{equation}


$\bullet$ HR-MAE: Computes the Mean Absolute Error (MAE) of HR between $\boldsymbol{g}$ and $\boldsymbol{d}$:
\begin{equation}
\text{HR-MAE} = |\mathrm{HR}(\boldsymbol{g}) - \mathrm{HR}(\boldsymbol{d})|,
\end{equation}
where $\mathrm{HR}(\cdot)$ is the function that derives the HR using the Heartpy package~\cite{van2019heartpy}.  

\subsection{Loss function}\label{ssec:loss}
We adopt a staged loss with a warmup strategy for the proposed DPNet. In the warmup epochs ($E_{\text{w}}$), the model is optimized with MSE ($\mathcal{L}_{\text{MSE}}$) and scale-invariant signal-to-distortion ratio (SI-SDR) loss ($\mathcal{L}_{\text{SI-SDR}}$)~\cite{8683855} to stabilize training and recover waveform fidelity. After $E_{\text{w}}$, an additional MAE loss ($\mathcal{L}_{\text{MAE}}$) between the GT and HRP–estimated BPM is introduced to encourage physiological consistency.
\begin{equation}
\mathcal{L} =
\begin{cases}
\mathcal{L}_{\text{MSE}} + \lambda_{1}\cdot\mathcal{L}_{\text{SI-SDR}}, & \text{if } E < E_{\text{w}}, \\
\mathcal{L}_{\text{MSE}} + \lambda_{1}\cdot\mathcal{L}_{\text{SI-SDR}} + \lambda_{2}\cdot\mathcal{L}_{\text{MAE}}, & \text{if } E \geq E_{\text{w}},
\end{cases}
\end{equation}
where $\lambda_{1}$ and $\lambda_{2}$ are the weights for balancing the magnitude of losses.

\subsection{Implementation details}
\label{ssec:implement}
The training procedure consists of two stages. First, the HRP is pre-trained for 200 epochs using the MSE loss to ensure reliable BPM estimation, achieving an HR-MAE of 1.014. In the second stage, the HRP is fixed while the DPNet is trained for 600 epochs with the warmup strategy described in Subsect.~\ref{ssec:loss}. We set $E_{\text{w}}=300$, $\lambda_{1}=10^{-4}$, and $\lambda_{2}=10^{-3}$. Both models are trained with the Adam optimizer~\cite{kinga2015method}, a learning rate of $10^{-5}$, and a batch size of 64. The best checkpoints are selected based on validation performance. 

\subsection{Results and discussion}
\label{ssec:result}
\begin{figure}[tbp]
    \centering
    \includegraphics[width=\columnwidth]{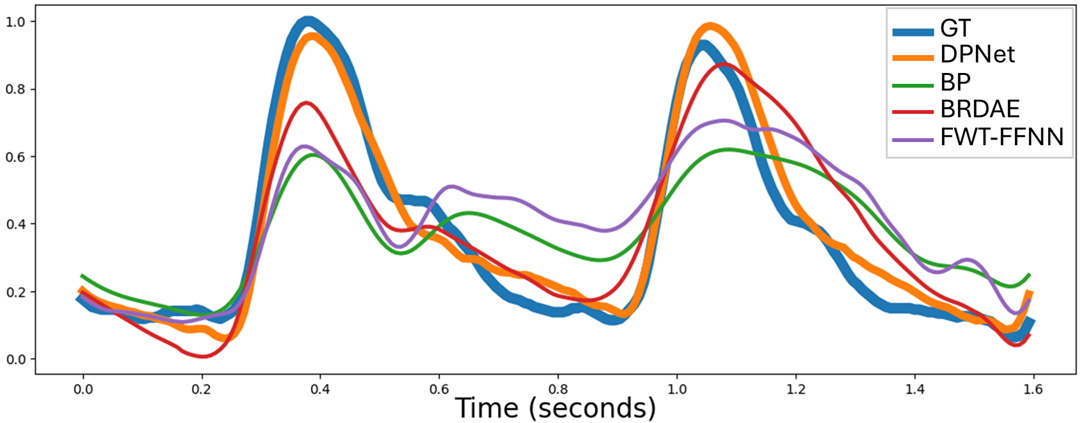}
    \caption{Denoised PPG signals using different methods.}
    \label{fig:result}
\end{figure}

Table~\ref{tab:comparison} summarizes the performance of our proposed framework, a bandpass (BP) filtering method~\cite{shenoi2005introduction}, BRDAE~\cite{lee2018bidirectional}, and FWT-FFNN~\cite{ahmed2023deep}.
Our method consistently outperforms all other approaches across every evaluation metric. Notably, the proposed DPNet achieves the lowest MSE of 6.663 and HR-MAE of 1.025, indicating the best denoising quality and superior preservation of HR information.
The BP method lacks the adaptability to address complex and non-stationary artifacts. As a result, it performs the worst across most metrics.
The BRDAE model, which adopts a two-layered BLSTM structure, could not fully capture long-range temporal dependencies, leading to limited denoising performance and suboptimal HR estimation.
Although FWT-FFNN combines CNN and wavelet transform to capture time- and frequency-domain features, it may disrupt the long-range temporal structure of PPG signals, which leads to the poorest performance.
Fig.~\ref{fig:result} provides a visual comparison of the denoised waveforms across different methods. Compared with other baselines, the proposed DPNet produces signals with better alignment with the GT.


Next, we explored the effect of incorporating different losses into the training objective. Table~\ref{tab:ablation_hrloss} shows the ablation study on different combinations of loss functions. Using only $\mathcal{L}_{\text{MSE}}$ provides a reasonable baseline, but the model mainly focuses on point-wise accuracy, resulting in suboptimal HR estimation.
Introducing an additional $\mathcal{L}_{\text{SI-SDR}}$ yields consistent improvements in both metrics, as it encourages the preservation of waveform shape and temporal alignment.
The most notable gain is achieved when the HR predictor is incorporated with an auxiliary $\mathcal{L}_{\text{MAE}}$. This directly guides the model to retain physiologically significant features for HR estimation, reducing HR-MAE by nearly 50\% compared to the $\mathcal{L}_{\text{MSE}}$ baseline and further decreasing the reconstruction error. These results highlight the complementary roles of the three objectives: MSE ensures point-wise fidelity, SI-SDR promotes structural preservation, and HR-aware supervision enforces physiological consistency, jointly leading to the best overall performance.

Finally, we validate the effectiveness of our DPNet by only replacing the BMamba blocks with alternative sequential models. As shown in Table~\ref{tab:ablation_mamba}, the BMamba achieves the best overall results, with both the lowest MSE (6.663) and HR-MAE (1.025), demonstrating its ability to capture long-range temporal dynamics while preserving physiologically meaningful structures. In contrast, the Transformer encoder performs the worst, yielding the highest errors and showing difficulty in modeling the repeating rhythmic patterns of PPG under limited data. The BLSTM attains better HR-MAE by capturing local dependencies but suffers from higher reconstruction error, reflecting weaker capacity in preserving waveform fidelity. These findings highlight the superiority of BMamba for PPG denoising compared with conventional Transformer and BLSTM architectures.

\begin{table}[tbp]
\centering
\renewcommand\arraystretch{1.1}
\small
\caption{Effect of different training loss used to train DPNet.}
\label{tab:ablation_hrloss}
\begin{tabular}{lcc}
\toprule
Loss combinations & MSE ($\times 10^{-3}$) $\downarrow$ & HR-MAE $\downarrow$\\
\midrule
$\mathcal{L}_{\text{MSE}}$ & 7.657 & 1.979 \\
$\mathcal{L}_{\text{MSE}}$ + $\mathcal{L}_{\text{SI-SDR}}$ & 7.255 & 1.856 \\
$\mathcal{L}_{\text{MSE}}$ + $\mathcal{L}_{\text{SI-SDR}}$ + $\mathcal{L}_{\text{MAE}}$ & \textbf{6.663} & \textbf{1.025} \\
\bottomrule
\multicolumn{3}{l}{\textbf{Bold} represents the best performance.}
\end{tabular}
\end{table}

\begin{table}[tbp]
\centering
\renewcommand\arraystretch{1.1}
\small
\caption{Comparison of Temporal Modeling Architectures.}
\label{tab:ablation_mamba}
\begin{tabular}{lcc}
\toprule
Block & MSE ($\times 10^{-3}$) $\downarrow$ &  HR-MAE $\downarrow$ \\
\midrule
Transformer & 10.017 & 4.862 \\
BLSTM & 9.828 & 1.149 \\
DPNet (BMamba) & \textbf{6.663} & \textbf{1.025} \\
\bottomrule
\multicolumn{3}{l}{\textbf{Bold} represents the best performance.}
\end{tabular}
\end{table}

\section{Conclusion}

In this study, we present a novel PPG denoising framework that leverages the Mamba architecture to effectively capture critical temporal features. To the best of our knowledge, this is the first work in PPG denoising to incorporate downstream-specific supervision. Experimental results also show that the auxiliary SI-SDR loss further enhances performance, while the HRP improves both waveform fidelity and physiological consistency significantly. Overall, DPNet consistently outperforms conventional filtering and existing neural approaches, achieving robust denoising against both synthetic noise and real-world motion artifacts. We consider the proposed DPNet can bridge the gap between signal processing performance and downstream clinical relevance, paving the way for broader adoption in wearable healthcare systems.


\vfill\pagebreak
\bibliographystyle{IEEEbib}
\bibliography{refs}

\end{document}